\documentclass[letterpaper, 10 pt, conference]{ieeeconf}  

\IEEEoverridecommandlockouts                              

\usepackage{graphics} 
\usepackage{epsfig} 
\usepackage{mathptmx} 
\usepackage{times} 
\usepackage{cite}
\usepackage[caption=false,font=footnotesize]{subfig}
\usepackage{amsmath,amssymb,amsfonts}
\usepackage{algorithmic}
\usepackage{graphicx}
\usepackage{textcomp}
\usepackage{hyperref}
\usepackage[T1]{fontenc}
\usepackage{color}
\def\BibTeX{{\rm B\kern-.05em{\sc i\kern-.025em b}\kern-.08em
    T\kern-.1667em\lower.7ex\hbox{E}\kern-.125emX}}

\title{\LARGE \bf
{Modeling Interactions of Multimodal Road Users in Shared Spaces}
}

\author{Fatema T. Johora$^{1}$ and J\"org P. M\"uller$^{2}$
\thanks{This research has been supported by the German Research Foundation (DFG) through the Research Training Group SocialCars: Cooperative (De-) centralized Traffic Management (GRK 1931). We cordially thank Federico Pascucci and Bernhard Friedrich for providing road user datasets.}%
\thanks{This paper has been published with copyright by IEEE, accessible at \url{https://ieeexplore.ieee.org/document/8569687}.}
}

\begin{document}

\maketitle
\thispagestyle{empty}
\pagestyle{empty}

\begin{abstract}

In shared spaces, motorized and non-motorized road users share the same space with equal priority. Their movements are not regulated by traffic rules, hence they interact more frequently to negotiate priority over the shared space. To estimate the safeness and efficiency of shared spaces, reproducing the traffic behavior in such traffic places is important. In this paper, we consider and combine different levels of interaction between pedestrians and cars in shared space environments. Our proposed model consists of three layers: a layer to plan trajectories of road users; a force-based modeling layer to reproduce free flow movement and simple interactions; and a game-theoretic decision layer to handle complex situations where road users need to make a decision over different alternatives. We validate our model by simulating scenarios involving various interactions between pedestrians and cars and also car-to-car interaction. The results indicate that simulated behaviors match observed behaviors well.
\end{abstract}

\section{INTRODUCTION}

In recent years, the shared space design paradigm introduced by Dutch traffic engineer Hans Monderman in the 1970s \cite{clarke2006shared}, has received considerable attention as an alternative approach
to classic traffic design. Shared spaces are mixed traffic environments, in which heterogeneous road users such as pedestrians, cyclists, and motorized vehicles are not separated by temporal and spatial division of road resources, but share the same space; their movements are conducted based on social protocols (e.g. courtesy behavior) and informal rules (e.g. colored road surface) instead of traffic regulations. They interact more frequently to coordinate trajectories, by performing actions they choose through negotiating with others.

Despite the ongoing debate on efficiency (average road user delays and road capacity) and safeness of shared space,
many cities in Europe already have adopted the shared space
design principles to reconstruct city centers \cite{schonauersocial}. This makes the investigation of traffic performance in shared spaces critical. Modeling and analyzing the motion behavior of
multimodal road users including interactions among them can
reproduce the operation of shared spaces, which provides a basis for  measuring aspects of their performance.

Modeling mixed traffic interactions is challenging as it is the result of complex human decision-making processes. To the best of our knowledge, until now, only little work has been done on modeling and simulating shared spaces. Previous works on shared spaces modeling use a physics-based model, most prominently the classical social force model (SFM) \cite{helbing1995social} to model pedestrian movement behaviors and extend SFM using long-range conflict avoidance mechanisms \cite{pascucci2015modeling} \cite{rinke2017multi}, rule-based constraints \cite{anvari2015modelling}, a multinomial logit model \cite{pascucci2018should}, or a game theoretic approach\cite{schonauer2012modeling} for modeling multimodal interaction. Cellular automata are also used to model mixed traffic in presence of traffic regulations like in \cite{lan2005inhomogeneous}, for modeling car-following and lane-changing behaviors of cars and motorcycles or in \cite{zhang2007modeling}, for modeling pedestrians-vehicles interactions at crosswalks.

All these works cover interactions among multimodal road users at a certain level of details. A restriction of the state of the art is that only single bilateral conflicts are considered. This means that, at a time, each road user can only handle a single conflict with another user, but not all conflicts she is encountering at that time. We define a conflict  in accordance with \cite{gettman2003surrogate} as ``an observable situation in which two or more road users approach each other in time and space to such an extent that there is a risk of collision, if their movements remain unchanged''. These models also do not consider courtesy behavior of car drivers for car-car interactions.

This paper describes a first step to address this gap, by proposing a multiagent-based model for more realistically describing, recognizing and dealing with real-world interactions of pedestrians and cars in shared spaces. It includes interactions involving two or more road users and the courtesy behavior of car drivers. 

\section{Problem Statement and Requirements}
\label{sec:Problem}
As there are no explicit traffic regulations in shared spaces, road users interact with each other for negotiating the priority of the integrated space.
To understand their interactions, we analyze the data of a shared space area in Hamburg \cite{pascucci2018should}.
Based on our observation of the video data and on the classification of road users' behaviors given by Helbing and Moln\`{a}r  \cite{helbing1995social}, we classify typical interactions into the following categories:
\begin{itemize}
	\item \textbf{Simple Interaction}: direct mapping of percepts to actions.
	\begin{itemize}
		\item \textbf{Reactive Interaction}: When road users do not have time to think or plan their actions, they behave re-actively. As an example, if a pedestrian suddenly jumps in front of the vehicle, to avoid a serious collision, the car driver will either break strongly or steer away immediately. 
		\item \textbf{Car Following} (car only): Even-though in shared space, there is no defined lane for cars, but empirical observation indicates that cars merge into assumed lanes and follow the car in front \cite{anvari2015modelling}.
	\end{itemize}
    \item  \textbf{Complex Interaction}: road users need to decide an action among different alternatives.
   \begin{itemize}
    	\item \textbf{Implicit Interaction between Two Road Users}: Each road user chooses her best action to avoid conflicts with another user by predicting their action. E.g., in reality, when a car comes with a moderate speed, a pedestrian may predict that if she continues to walk, the car driver will decelerate and let her go.
    	\item \textbf{Implicit Interaction among Multiple Road Users}: Sometimes road user needs to implicitly interact with multiple users at the same time as in pedestrian crossing, where pedestrian needs to interact with cars coming from both directions to make a precise decision to cross or not to cross the road.
    	\item \textbf{Explicit Interaction}: Road users also interact with other road users through communicating.
     \end{itemize}
\end{itemize}
Road users interact by performing some actions. \textbf{Courtesy behavior} is one of them, where a road user decelerates and eventually stops to give priority to others. We model this behavior both as reactive and complex interaction.

The \textbf{simple interactions} can be represented using forced-based \cite{helbing1995social, anvari2015modelling} or cellular automata models \cite{burstedde2001simulation, nagel1992cellular}. However, for \textbf{complex interactions}, decision-theoretic models such as games are useful\cite{helbing1995social}.  
In a noncooperative game-theoretic model, each decision maker makes a decision by predicting others' decision. 
We argue that such models can more precisely model \textit{implicit} interaction between road users. Indeed, they have been used to describe  merging-give way interaction \cite{kita1999merging}, lane-changing interaction of cars \cite{lutteken2016using}, or cyclist-car driver interaction at zebra crossings \cite{bjornskau2017zebra}. We note that modeling \textit{explicit} interaction is also missing in previous works. However, we do not consider it in this paper.

\section{Multiagent-Based Simulation Model}
\label{Proposed model}
\begin{figure}[htbp]
	\centering
	\includegraphics[width=3.5in]{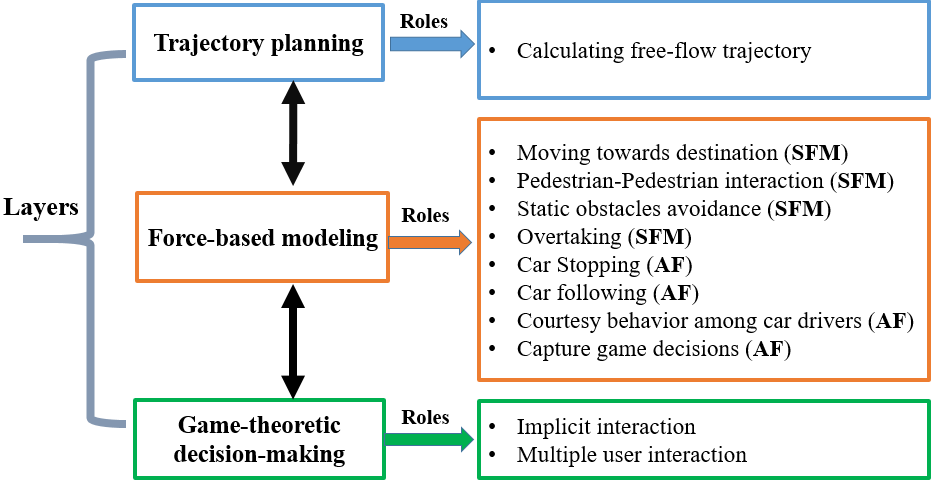}
	\caption{The conceptual structure of motion behaviors of multimodal road users. Here, AF means Additional Force and SFM means Social Force Model (adapted from \cite{anvari2015modelling},p.88).}
	\label{Transition Handler}
	\vspace{-0.5em}
\end{figure}

We model the motion behaviors of multimodal road users, including the interactions among them (discussed in Section \ref{sec:Problem}) in three layers: trajectory planning, force-based modeling, and game-theoretic decision-making. Figure \ref{Transition Handler} summarizes the roles of each layer. 

\subsection{Trajectory Planning Layer}

We plan the trajectories of road users by considering static obstacles like walls or trees in the shared space environment. To perform a trajectory planning algorithm, the environment needs to be represented as a graph. We transform the environment into a visibility graph: where each edge connects two outline vertices of any obstacle, which are visible to each other and these edges do not collide with any line segment
that outlines any obstacle\cite{koefoed2012representations}. The origin and destination points of each road user are added to the graph. We perform the A* algorithm\cite{millington2009artificial} on the graph to plan free flow trajectory for each road user. However, humans keep some distance from obstacles. Hence, to model this behavior we fine-tune the placement of the inner path vertices regarding their distance to obstacles.

\subsection{Force-Based Modeling Layer}
We use the classical social force model (see also \cite{Johora+2017abmus}) to model the driving force of road users towards their destination, the interaction of road users with static obstacles, the interaction between pedestrians, and also the overtaking behavior of cars. 
In this model, the motion of a road user is conducted by a set of simple forces that represent various influences she faces in simple situations: 
$$
\frac{ d\vec{v}_\alpha(t)}{ dt} = \vec{f}_{\alpha}^o  + \sum\vec{f}_{\alpha B}^{soc} + \sum_ {\beta \neq V } \vec{f}_{\alpha V}^{soc}+ \color{blue} \vec{I}_{\alpha\beta}^{stopping}+ \color{blue} \vec{I}_{\alpha\beta}^{courtesy} + \vec{I}_{\alpha\beta}^{following} + \vec{I}_{\alpha\beta}^{game}
$$
Here, $\vec{f}_{\alpha}^o$ represents the driving force of road user $\alpha$ and can be computed by the  formula:
$\vec{f}_{\alpha}^o(\vec{v^*}_\alpha(t), \vec{v}_\alpha(t) ) = \frac{\vec{v^*}_\alpha(t) - \vec{v}_\alpha(t)}{\tau}$ for a relaxation time $\tau$ and $\vec{v^*}_\alpha(t)$ and $\vec{v}_\alpha(t)$ are the desired and current velocities of $\alpha$ respectively.
Forces $\vec{f}_{\alpha\beta}^{soc}$ and $\vec{f}_{\alpha B}^{soc}$ represent the human behavior of maintaining some distance from other road users $\beta$ and static obstacles B. If $\alpha$ is a pedestrian then $\vec{f}_{\alpha\beta}^{soc}$ applies between $\alpha$ and every other road users, but if $\alpha$ is a car then $\vec{f}_{\alpha\beta}^{soc}$ only applies between $\alpha$ and every other cars. These behaviors are formulated by the following equations:
$f_{\alpha\beta}^{soc}(\vec{d}_{\alpha\beta}(t)) = V_{\alpha\beta}^o\exp\bigg[\frac{- \vec{d}_{\alpha\beta}(t)}\sigma\bigg] \hat{n}_{\alpha\beta} F_{\alpha\beta}$ 
and 
$\vec{f}_{\alpha B}^{soc}(\vec{d}_{\alpha B}(t)) =  U_{\alpha B}^o\exp\bigg[\frac{- \vec{d}_{\alpha B}(t)}{R}\bigg] \hat{n}_{\alpha B}$, where $V_{\alpha\beta}^o$ and $U_{\alpha B}^o$ denote the interaction strengths, and $\sigma$ and $R$ indicate the range of these repulsive interactions. $\vec{d}_{\alpha\beta}(t)$ and $\vec{d}_{\alpha B}(t)$ are the distances from $\alpha$ to $\beta$, or $\alpha$ to $B$ at a specific time. $\hat{n}_{\alpha\beta}$ and $\hat{n}_{\alpha B}$ denote the normalized vectors. $F_{\alpha\beta}$ represents the fact that human are mostly influenced by the objects which can be captured within their field of view.

We extend the social force model to consider reactive interactions of cars towards pedestrians ($\color{blue} \vec{I}_{\alpha\beta}^{stopping}$), 
car following interactions ($\color{blue} \vec{I}_{\alpha\beta}^{following}$) and courtesy behavior of car drivers ($\color{blue} \vec{I}_{\alpha\beta}^{courtesy}$):
\paragraph{Car to Pedestrian Reactive Interaction:}
If the nearest conflict of a car driver $\alpha$ is with pedestrian(s) $\beta$ and $\beta$ has already started moving in front of $\alpha$ or is very close to $\alpha$, then $\alpha$ decelerates to let $\beta$ pass. This type of interaction only happens if $\theta_{\vec{e}_\alpha\hat{n}_{\beta\alpha}}$ $\leq$ 15 $\parallel$ $\theta_{\vec{e}_\alpha\hat{n}_{\beta\alpha}}$ $\geq$ 345.
 \paragraph{Car Following}
 \label{carfollowing}If $d_{\alpha\beta}$ $\geq$ $d_\alpha$, the car following interaction of follower $\alpha$ with leader $\beta$ is $\vec{I}_{\alpha\beta}^{following}$ = $\hat{n}_{p_\alpha x_\alpha}$, $\alpha$ maintains the direction of movement towards $p_\alpha = \vec{x}_\alpha(t) + \hat{v}_\beta(t) \cdot d_\alpha$, where $d_\alpha$ is the minimum vehicle distance, $d_{\alpha\beta}$ is the distance between $\alpha$ and $\beta$, and $\vec{x}_\alpha(t)$ is the current position of $\alpha$. Otherwise, $\vec{I}_{\alpha\beta}^{following}$ = $\frac{s_{current}}{2}$, $\alpha$ slows down to maintain safe distance from $\beta$. This interaction type only occurs if the leader $\beta$ is in the movement direction of the follower $\alpha$ ($\vec{e}_\alpha$): $\theta_{\vec{e}_\alpha\hat{n}_{\beta\alpha}}$ $\leq$ 10 $\parallel$ $\theta_{\vec{e}_\alpha\hat{n}_{\beta\alpha}}$ $\geq$ 350 and the movement direction of $\beta$ ($\vec{e}_\beta$) is $\sim$ equal to $\vec{e}_\alpha$: $\theta_{\vec{e}_\beta\hat{n}_{\beta\alpha}}$ $\leq$ 5 $\parallel$ $\theta_{\vec{e}_\alpha\hat{n}_{\beta\alpha}}$ $\geq$ 355. Here, $\hat{n}_{\beta\alpha}$ is the normalized vector from $\beta$ to $\alpha$ and $\theta_{\vec{e}_\beta\hat{n}_{\beta\alpha}}$ is the angle between vectors $\vec{e}_\alpha$ and $\hat{n}_{\beta\alpha}$. The force $f_{\alpha\beta}^{soc}$ is set to zero when car following is active.
\paragraph{Courtesy Behavior}
\label{courtesy}
The courtesy behavior of driver $\alpha$ to another road user $\beta$ is $\vec{I}_{\alpha\beta}^{courtesy}$ = $\frac{s_{current}}{2}$, here $\alpha$ decelerates and eventually stops when $\beta$ is passing in front of $\alpha$. This type of interaction only happens if $\theta_{\vec{e}_\alpha\hat{n}_{\beta\alpha}}$ $\leq$ 20 $\parallel$ $\theta_{\vec{e}_\alpha\hat{n}_{\beta\alpha}}$ $\geq$ 340 and 70 $<$ $\theta_{\vec{e}_\beta\hat{n}_{\beta\alpha}}$ $<$ 100 $\parallel$ 260 $<$ $\theta_{\vec{e}_\beta\hat{n}_{\beta\alpha}}$ $<$ 290.
 The force $f_{\alpha\beta}^{soc}$ is set to zero when courtesy behavior of $\alpha$ is active.

This layer is also responsible for executing the game layer decisions.
As an example, if the result of a game-theoretic interaction between a car driver and a pedestrian is that the pedestrian will continue to walk and the vehicle driver will decelerate (see also Subsection~\ref{subsec:game}), the respective actions of car and pedestrian will be executed in the Force-Based Modeling Layer. In this paper, we consider the following possible actions for road users: 
\begin{itemize}
	\item Continue: Car drivers continue the free-flow movement with their previous speed. Pedestrians also continue free-flow movement but with maximum speed.
	\item Decelerate: Road users decelerate and eventually stop.
	\item Deviate (pedestrian only): We model this action as that any pedestrian $\alpha$ first passes the car $\beta$ from behind from a position $p_\alpha$ and then continues moving towards the original destination. Here,
	$p_\alpha = \vec{x}_\beta(t)$ - 0.5$\vec{v}_\beta(t)$, if the car continues moving and
	$p_\alpha = \vec{x}_\beta(t)$ - $\vec{v}_\beta(t)$, if the car stops.
	The pedestrian will move towards $p_\alpha$ until the car stays within the field of view (FOV) of the pedestrian.
\end{itemize}

\subsection{Game-Theoretic Decision Layer}\label{subsec:game} This layer handles complex interactions among road users. We use one-shot Stackelberg games, which are sequential leader-follower games, where the leader player commits to an action first; then followers choose an action to optimize their utility considering the action chosen by the leader \cite{schonauer2012modeling}. 
For a single game, restrict the number of leaders to one, while there may be one or more followers. For each implicit interaction only one game is played and the games are independent of each other. In multiple players game, the leader player makes a decision $s_l$ by considering the nearest follower and the total number of followers. All followers follow the same action $s_f$.

Solving the Stackelberg game is done by finding the subgame perfect Nash equilibrium (SPNE), which can be inferred by backward induction. Let $S_l$ and $S_f$ are the sets of all possible actions of the leader and follower players respectively, and $u_l(s_l, s_f)$ and $u_f(s_f|s_l)$ are the utility of the leader and followers, when the leader choose the action $s_l \in S_l$ and the followers react by the move $s_f \in S_f$. 
$\forall$$s_l\in S_l$, $B_f(s_l)$ = \{$s_f\in S_f|max(u_f(s_f|s_l))$\} is the best answer of the follower.
Then the SPNE = $<$\{$s_l\in S_l | max(u_l(s_l, B_f(s_l)))$\}, $B_f(s_l)$$>$.
 
\subsubsection{Payoff Estimation}
\begin{figure}[htbp]
	\centering
	\subfloat[Ordinal valuations of outcomes.]{\includegraphics[width=1.4in]{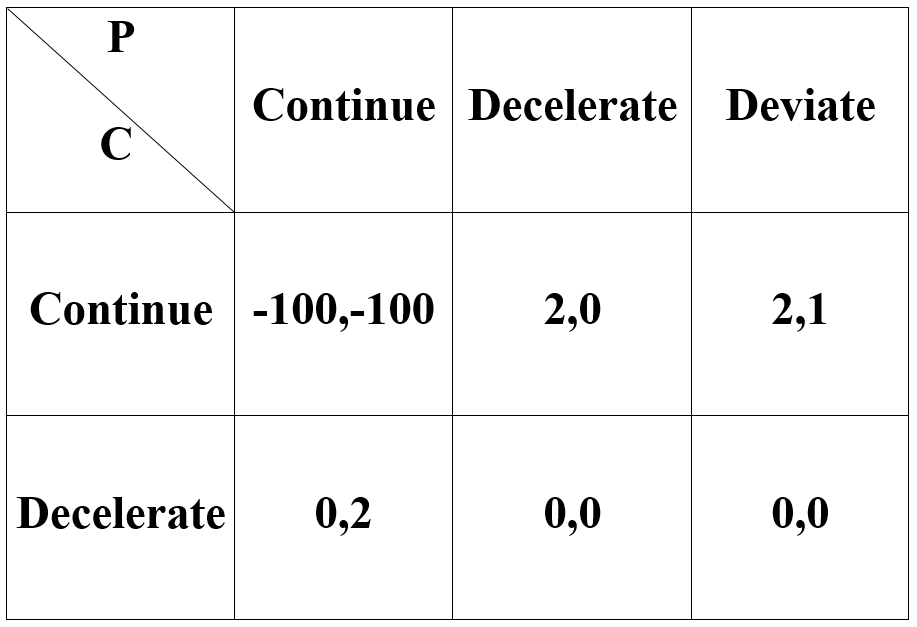}
		\label{ordinal}}
	\hfil
	\subfloat[The complete payoff matrix]{\includegraphics[width=1.33in]{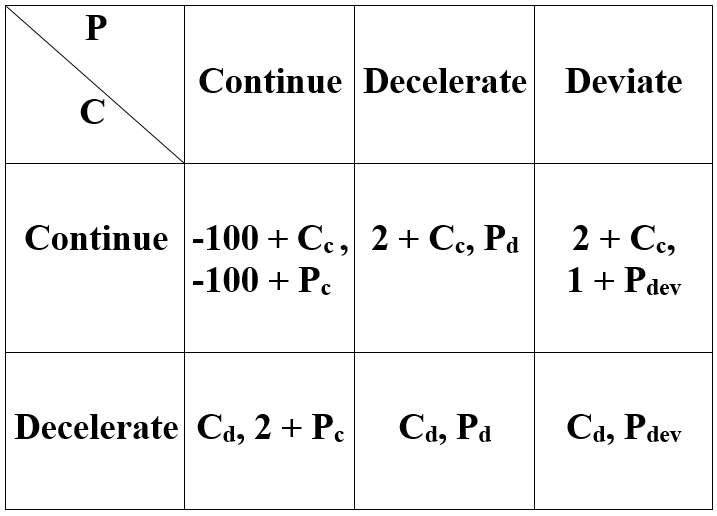}
		\label{with_parameter}}
	\caption{The row player is a pedestrian with three possible actions; the column player is a car driver with two possible actions.}
	\label{game}
\end{figure}

In the game, all actions of the players are ordinally valued from 2 to -100 as shown in Figure \ref{ordinal}, with the assumption that every person prefers to reach their destination safely and quickly. To model courtesy behavior and situation dynamics, we consider six observable factors (in the following denoted by Boolean variables  $x_1 \ldots x_6$) relevant for payoff estimation. They are defined in the following and are used to calculate a set of parameters $F_1 \ldots F_{12}$, which are impacts of these factors. Let $\alpha$ is a road user who needs to interact with another user $\beta$:
\begin{itemize}
	\item $x_1$: has value $1$, if current speed of $\beta$, $S_{current}$ $<$ $S_{normal}$. This factor determines the value of $F_1$, $F_6$ and $F_{11}$.
	\item $x_2$: has value $1$ if number of active interactions $<$ N (=2). This factor determines the value of $F_2$ and $F_7$.
	\item $x_3$: has value $1$ if the number of give ways already allowed by $\alpha$, $G_\alpha$ $<$ M (=3) or not. This determines the value of $F_3$ and $F_8$. We generate the value of $G_\alpha$ randomly.
	\item $x_4$: has value $1$ if  $\alpha$ is a car driver following another car. This factor determines the value of $F_4$ and $F_9$.
	\item $x_5$: has value $1$ if  $\alpha$ is a pedestrian in a group. This factor determines the value of $F_5$ and $F_{10}$.
	\item $x_6$: has value $1$ if $\alpha$ is a car driver followed by another car. This factor determines the value of $F_{12}$.
\end{itemize}

For the scope of this paper, the parameters N and M were chosen in a ``trial-and-error'' manner through simulation experiments. \\
The payoff of a car driver's actions is calculated as follows:
\begin{itemize}
\item Continue: $C_c$ = $F_1$ + $F_2$ + $F_3$ + $F_4$ + $F_5$
\item Decelerate: $C_d$ = $F_6$ + $F_7$ + $F_8$ + $F_9$ + $F_{10}$
\end{itemize}
The payoff value of the actions of a pedestrian is calculated as follows:
\begin{itemize}
	\item Continue: $P_c$ = $F_1$ + $F_9$ + $F_{10}$
	\item Decelerate: $P_d$ = $F_6$ + $F_4$ + $F_5$
	\item Deviate: $P_{dev}$ = $F_{11}$ + $F_{12}$
\end{itemize}

The utility calculation of the action pairs is shown in Figure \ref{with_parameter}. 
As we were not able to determine the parameters $F_1$, $F_2$, $F_3$, $F_4$, $F_5$, $F_6$, $F_7$, $F_8$, $F_9$, $F_{10}$, $F_{11}$, and $F_{12}$ based on a substantial amount of real interaction scenarios, there remains a considerable doubt about providing correct values for other and more complex scenarios. Hence, while we did perform a sensitivity analysis on a certain amount of interaction scenarios and analyzed the behavior of our model, we do not present the values of these parameters in this paper. Automated calibration and validation of these parameters (including N and M) is an important task in our future work, see Section~\ref{sec:conclusion}.

\subsection{Recognition and Handling of Interactions}
In our model, simple interactions are handled locally by the road users. Remember that simple interactions are dealt with by local action and does not require synchronization or communication. If a simple interaction situation is detected by a road user within its FOV (set to 180{\textdegree} in this paper), it is classified based on the interaction angle as given in Section \ref{sec:Problem}, and the corresponding action is performed. Complex interactions are handled by a (virtual) central mediation entity, further referred to as \textit{host agent}. At first, the host agent collects the nearest conflict (one conflict at a time) and the total number of conflicts the user agent is facing currently for each car agent in the environment by considering their FOV. The host agent checks if this conflict can be solved without game playing, if yes then the conflict will be handled by the respective car agent itself. Otherwise, the host selects the car agent (faster agent) as the leader and
organizes a game by considering only the selected leader and its followers, not all other user agent involved in this situation (if any), as then the host may need to consider all possible subsets of conflicts as one game (worst-case scenario) which will make the situation more complex. Hence, the impact of follower-to-follower conflict and follower-to-other-agent conflict is not considered in the scope of a single game. The workflow of a user agent and the host agent for a single cycle are shown in Figure \ref{fig_agent}. The agent cycle is executed once per simulation time unit including time to perform all interactions.

\begin{figure}[htbp]
	\centering
		\vspace{-0.3em}
	\subfloat[ A Road User Agent]{\includegraphics[width=2.6in]{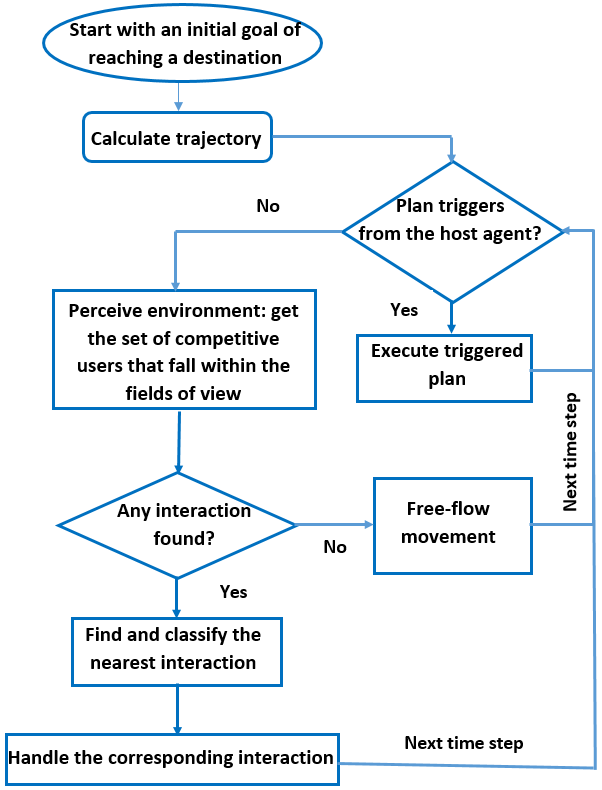}
		\label{user_agent}}
	\hfil
		
	\subfloat[ The Host Agent ]{\includegraphics[width=2.5in]{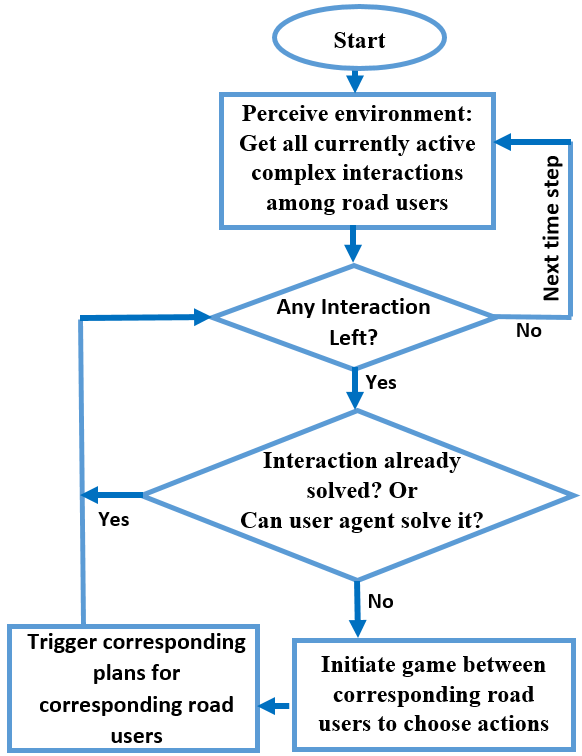}
		\label{host_agent}}
	\caption{The inner steps of the user and host agents with their plans and goals.}
		\vspace{-1em}
	\label{fig_agent}
\end{figure}

\section{Evaluation}\label{evalution} 
To evaluate our simulation model, we simulate 117 conflict scenarios from the shared space data and visualize the real and simulated trajectories and speed difference of all road users involved in these scenarios. We select two example scenarios among all those scenarios that involve multiple user interaction, the courtesy behavior of car drivers towards pedestrians and also the car following behavior to illustrate elaborately how our model works. 
We also simulate another scenario revealing courtesy behavior of car drivers towards other car drivers. However, for the last case we could not compare the performance as real data is currently missing. 
We consider all these three scenarios as complex situations because here the road users can choose their actions among different alternatives like the pedestrians can deviate, continue, or stop moving and the car driver can give way or continue driving. We did not consider deviation as an option for car drivers because of the non-holonomic constraints of the vehicular movement. To simulate different shared space scenarios, we use Lightjason, a Java-based BDI multi-agent framework \cite{aschermann2016lightjason}. All simulation runs were performed on an Intel$\circledR$ Core\texttrademark i5 processor with 16 GB RAM.
\paragraph{Scenario 1} Here, an interaction between three pedestrians and one car driver is captured. In this scenario, the car driver decelerates to let the pedestrians cross the street.
In our model, car handles this interaction re-actively as the angle between the car and (at least) one of the pedestrians is $\leq$ 15 $\parallel$ $\geq$ 345 and reproduces the same solution.
 \begin{figure}[htbp]
\centering
\subfloat[observation]{\includegraphics[width=2.75in]{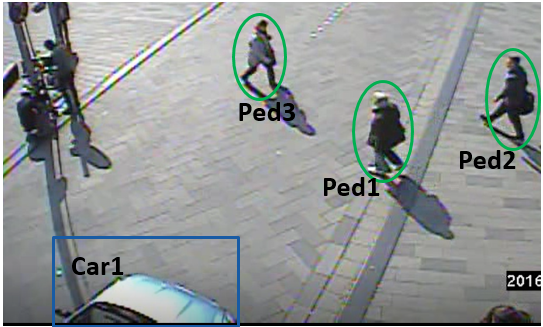}
\label{scenario1_observation}}
\hfil
\vspace{-0.05em}
\subfloat[Comparison of the intensity of interaction]{\includegraphics[width=3in]{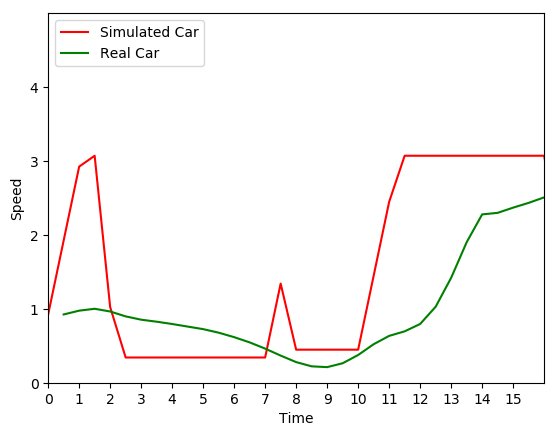}
\label{speeddif}}
\hfil
\subfloat[Comparison of the trajectories. The arrows indicate the direction of movement of the road users.]{\includegraphics[width=3.4in]{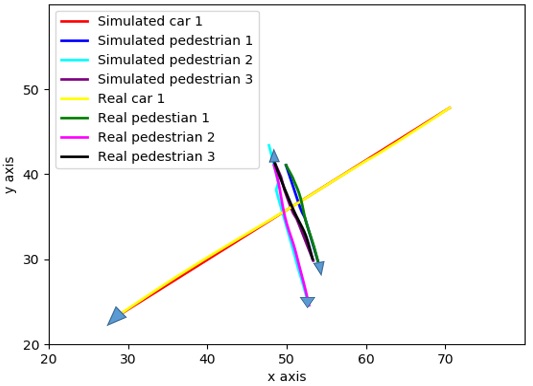}
	\label{withgame}}
\caption{Comparison of a car-to-pedestrian interaction}
\vspace{-2em}
\label{fig4}
\end{figure}
We observe that in both simulation and reality, the car driver decelerates; yet, the timing and the way of acting differ in both cases: The acceleration and deceleration of simulated car driver is sharp compare to real car driver. The mechanism of speed handing of cars in our model need to be improved.
The speed difference between the real and simulated cars are illustrated in Figure \ref{speeddif}.
The comparison between the observed and simulated trajectories is given in Figure \ref{withgame}. In Figure \ref{withgame}, the trajectories of both simulated and real car are very similar, while the trajectories of pedestrians in simulation slightly deviated from the real trajectories because of the repulsive force between pedestrians when they come close to each other.  
\begin{figure}[htbp]
	\centering
    \subfloat[]{\includegraphics[width=1.50in]{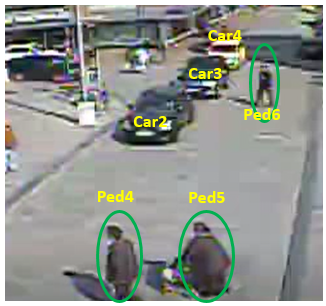}%
		\label{fig_scenario2first}}
	\hfil
	\vspace{-0.01em}
	\subfloat[The game matrices, left: \textbf{Car2} to \textbf{Ped4} and \textbf{Ped5}, right: \textbf{Car4} to \textbf{Ped6} ]{\includegraphics[width=3.4in]{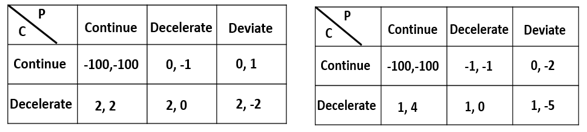}
		\label{fig_scenario2matrix}}	
	\hfil
	\vspace{-0.01em}
	\subfloat[The real trajectories ]{\includegraphics[width=2.7in]{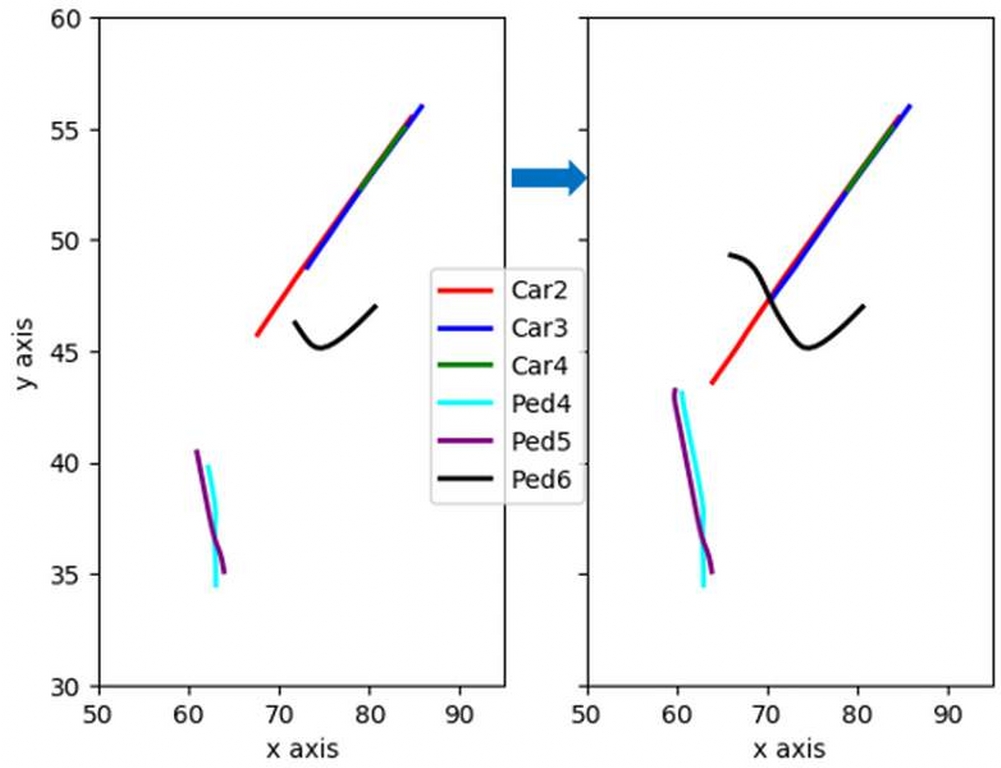}
		\label{fig_scenario2third}}
	\hfil
	\vspace{-0.01em}
	\subfloat[The simulated trajectories]{\includegraphics[width=2.7in]{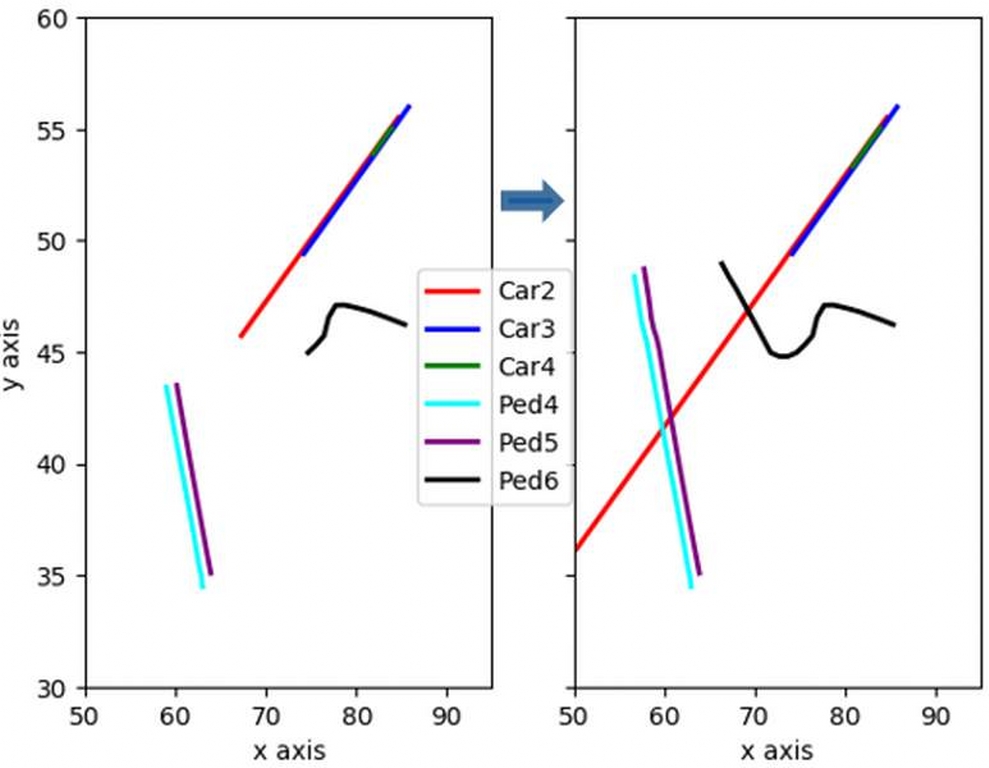}
		\label{fig_scenario2forth}}
	\hfil
    \vspace{-0.15em}
    \subfloat[The simulated trajectories without game decision]{\includegraphics[width=2.7in]{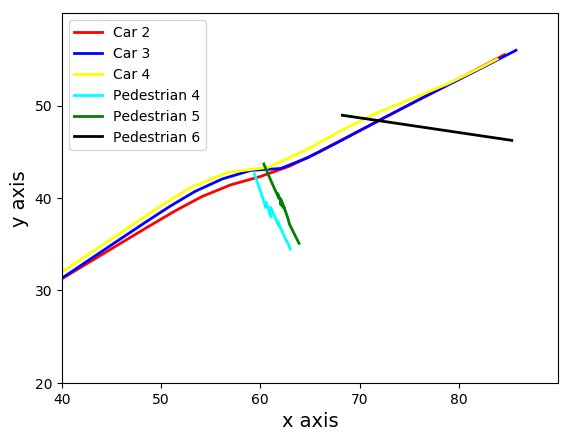}
    	\label{fig_scenario2fifth}}
	\caption{Comparison of a car to pedestrian and a car to car interaction}
	\vspace{-1.4em}
	\label{scenario2}
\end{figure}

\paragraph{Scenario 2}
It includes two types of interactions, namely car-to-pedestrian and car-to-car. In this scenario, \textbf{Car2} has interaction with \textbf{Ped4} and \textbf{Ped5}; \textbf{Car3} and \textbf{Car4} have car following interaction with \textbf{Car2}; and \textbf{Car4} has interaction with \textbf{Ped6}. Here, both \textbf{Car2} and \textbf{Car4} decelerate to let the pedestrians cross the street. We handle both car-to-pedestrian interactions using game-theoretic model and depict the game matrices in Figure \ref{fig_scenario2matrix}. Here, both \textbf{Car2} and \textbf{Car4} act as leader and decide to decelerate; their payoff function reflects courtesy of the drivers towards the pedestrians, as decelerating will maximize its value. 
We visualize the real and simulated trajectories of these interactions at two specific time steps in Figure \ref{fig_scenario2third} and Figure \ref{fig_scenario2forth} respectively. In both Figure \ref{fig_scenario2third} and Figure \ref{fig_scenario2forth}, \textbf{Ped} indicates Pedestrian. 
\textbf{Car3} and \textbf{Car4} in Figure \ref{fig_scenario2forth} decide to stay behind \textbf{Car2}, instead of overtaking as in Figure \ref{fig_scenario2third}. We model this behavior as reactive car following. Our simulation model reproduces real interactions, but a large safety distance is maintained between pedestrian to pedestrian, car to car and also car to pedestrian, compared to real situation. 

We simulate this same scenario again but without game-theoretic layer. Figure \ref{fig_scenario2fifth} visualizes the simulated trajectories at a specific time step to show how the classical social force model behaves in this situation. In Figure \ref{fig_scenario2fifth} all cars deviate from their trajectory instead of stopping and pedestrians sometimes go back and forth and sometimes stop. 

\begin{figure}[htbp]
	\centering
	\includegraphics[width=3.2in]{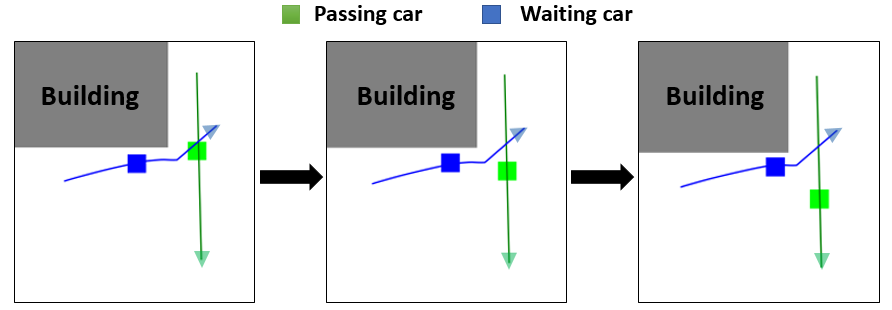}
	\caption{Car-to-car interaction at different time steps. The blue car is showing courtesy to the green car.}
	\label{scenario3}
	\vspace{-0.35em}
\end{figure}

\paragraph{Scenario 3}  The third example studies courtesy behavior of a car driver towards another. This is shown in Figure \ref{scenario3}. In the example, when the driver of the blue car perceives the green car, she slows down and lets the green car pass first.

To sum up, in Scenario 1, our simulation model reproduces a multiple user interaction among four road users realistically. Scenario 2 shows that our model can handle three interdependent situations realistically. In the third scenario, our model simulates the courtesy behavior between car drivers re-actively. In all these example scenarios, the road users suitably detect and classify the interaction and behave accordingly during simulation.

\begin{figure}[htbp]
	\centering
		\vspace{-0.5em}
	\subfloat[ The real trajectories]{\includegraphics[width=3.7in]{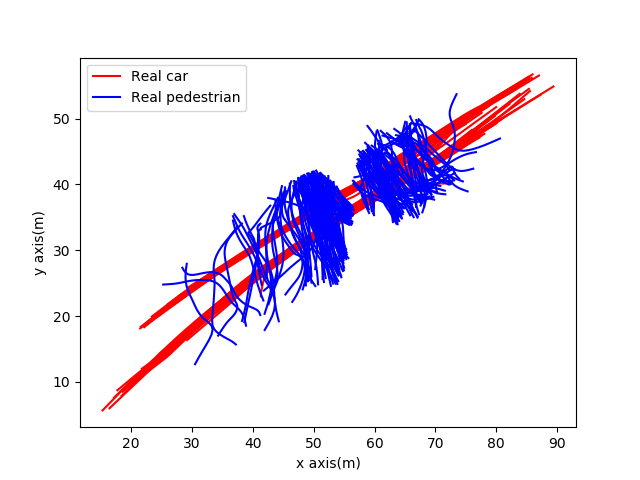}
		\label{realData}}
	\hfil
	\vspace{-.25em}	
	\subfloat[ The simulated trajectories ]{\includegraphics[width=3.7in]{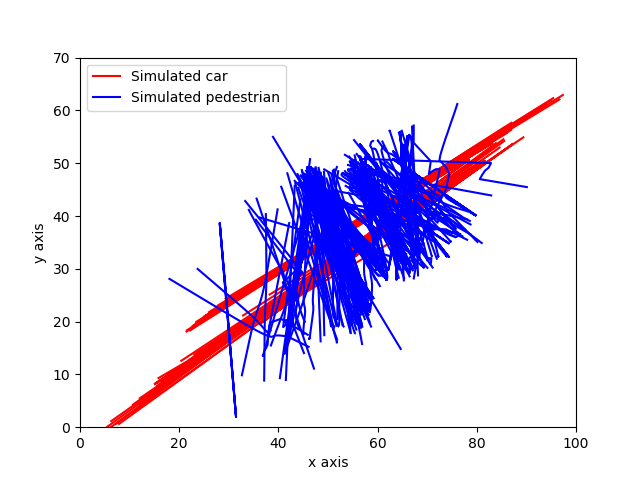}
		\label{simData}}
	\caption{The trajectories of pedestrians and cars are indicated by blue lines and red lines respectively.}
		\vspace{-1em}
	\label{fig_agent}
\end{figure}
\begin{figure}[htbp]
	\centering
    \subfloat[Distance between real and simulated trajectories]{\includegraphics[width=2.8in]{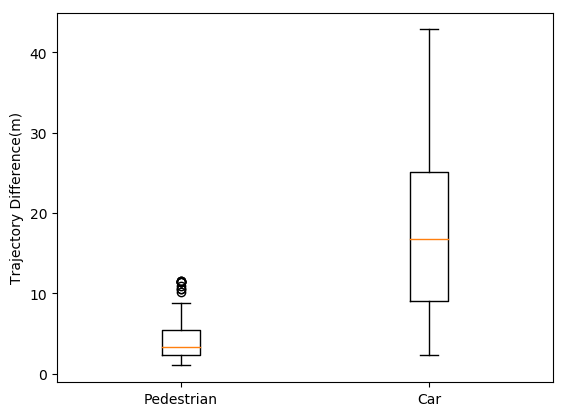}
		\label{fig_realcarspeed}}
	\hfil
	\subfloat[Speed differences of real and simulated road users]{\includegraphics[width=2.8in]{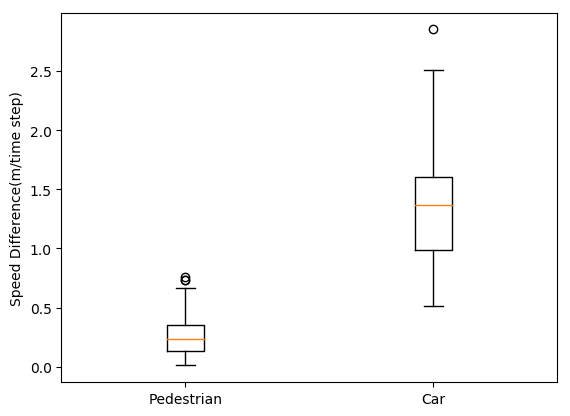}
		\label{fig_realpedspeed}}
			\caption{The difference in trajectory and speed of real and simulated pedestrians and cars.}
	\label{allscenario}
\end{figure}

Figure \ref{realData} and Figure \ref{simData} visualize the real and simulated trajectories of all road users involved in 117 selected real scenarios, respectively. We start simulating each road user five time steps before and end simulating five time steps after than real data, because in some situations road users cross each other without having any conflict, which is not the case in real scenario. The reason behind that is the speed of road users varies in real and simulated scenarios. The trajectories of real and simulated road users closely match. However, the time step wise position of each road user differs in real scenario and simulation because of speed difference between real and simulated road users and also the fact that a large safety distance is maintained between road users in simulation. The mean and standard deviation of speed of the simulated and real road users are: 2.779 and 2.966 (simulated cars); 0.8469 and 0.7813 (simulated pedestrians); 1.27 and 1.007 (real cars), 0.502 and 0.2526 (real pedestrians). The calibration of the acceleration and deceleration parameters is a part of our future work. Figure \ref{fig_realcarspeed} represents average distance between simulated and real trajectories of every car and pedestrian involved in each scenario of our selected 117 scenarios. For more than one car or pedestrian in a single scenario we take the average among them to calculate the average distance. Figure \ref{fig_realpedspeed} visualizes the individual average speed difference of real and simulated road users, which is calculated in the same way as trajectory distance. 

\section{Conclusion and Outlook}
\label{sec:conclusion}
This work-in-progress paper proposes a multiagent-based model for modeling interactions of pedestrians and cars in shared spaces. We analyze mixed traffic interactions, classify the interactions in terms of complexity and then model different levels of interactions using the social force model and Stackelberg games. Our model facilitates interaction recognition depending on the situation context. First results support the hypothesis that this model has good potential to simulate both complex and simple shared space scenarios, as modeling multiple user interactions and interdependent interactions can make the simulation model more realistic.

Future research will focus on calibrating and validating the model parameters using genetic algorithms. Next, we will try to find out other factors which might influence road user’s decision making. The explicit interaction and pedestrian group dynamic will be integrated into our simulation model. Most importantly, we shall study larger scenarios with larger numbers interactions. Doing so, we shall explore the scalability of various interaction types for different scenarios varying in number and heterogeneity of users as well as in topology. In particular, more efficient algorithms for recognizing, framing and resolving conflicts interactions will be required for realistically simulating larger shared space scenarios.

\bibliographystyle{IEEEtran}
\bibliography{references}

\end{document}